\documentclass[11pt]{article}

\usepackage[final]{acl}

\usepackage{times}
\usepackage{latexsym}

\usepackage[T1]{fontenc}

\usepackage[utf8]{inputenc}

\usepackage{microtype}

\usepackage{inconsolata}

\PassOptionsToPackage{table}{xcolor} 

\usepackage{graphicx}       
\usepackage{amsmath,amssymb,mathtools}
\usepackage{pifont}         
\usepackage{arydshln}       
\usepackage{booktabs}       
\usepackage{multirow}
\usepackage{algorithm}
\usepackage{algorithmic}
\usepackage[table]{xcolor}  
\usepackage{colortbl}

\usepackage{tikz} 
\usepackage{placeins}
\usepackage{caption}

\title{
	Bridging Reasoning and Action: Hybrid LLM–RL Framework for \\ Efficient Cross-Domain Task-Oriented Dialogue}

\author{
  \textbf{Yangyang Zhao\textsuperscript{1}},
  \textbf{Lifan Dai\textsuperscript{2}},
  \textbf{Li Cai\textsuperscript{5}},
  \textbf{Bowen Xing\textsuperscript{4}},
   \textbf{Libo Qin\textsuperscript{3, 5}} \thanks{corresponding author}
\\
\textsuperscript{1}School of Computer Science and Technology, \\  Changsha University of Science and Technology \\
\textsuperscript{2}School of Computer Science and Engineering, Central South University\\
\textsuperscript{3}Institute of Computing and Intelligence, Harbin Institute of Technology, Shenzhen\\
\textsuperscript{4}University of Science and Technology Beijing\\
\textsuperscript{5} Text Computing and Cognitive Intelligence MOE Engineering Research Center, \\
Guizhou University
}

\begin{document}
\maketitle
\begin{abstract}

Cross-domain task-oriented dialogue requires reasoning over implicit and explicit feasibility constraints while planning long-horizon, multi-turn actions. Large language models (LLMs) can infer such constraints but are unreliable over long horizons, while Reinforcement learning (RL) optimizes long-horizon behavior yet cannot recover constraints from raw dialogue. Naively coupling LLMs with RL is therefore brittle: unverified or unstructured LLM outputs can corrupt state representations and misguide policy learning. Motivated by this, we propose \textbf{V}erified \textbf{L}LM-\textbf{K}nowledge empowered \textbf{RL} (\textbf{VLK-RL}), a hybrid framework that makes LLM-derived constraint reasoning usable for RL. VLK-RL first elicits candidate constraints with an LLM and then verifies them via a dual-role cross-examination procedure to suppress hallucinations and cross-turn inconsistencies. The verified constraints are mapped into ontology-aligned slot–value representations, yielding a structured, constraint-aware state for RL policy optimization. Experiments across multiple benchmarks demonstrate that VLK-RL significantly improves generalization and robustness, outperforming strong single-model baselines on long-horizon tasks.

\end{abstract}

\section{Introduction}

\begin{figure}[t]
	\includegraphics[width=\columnwidth]{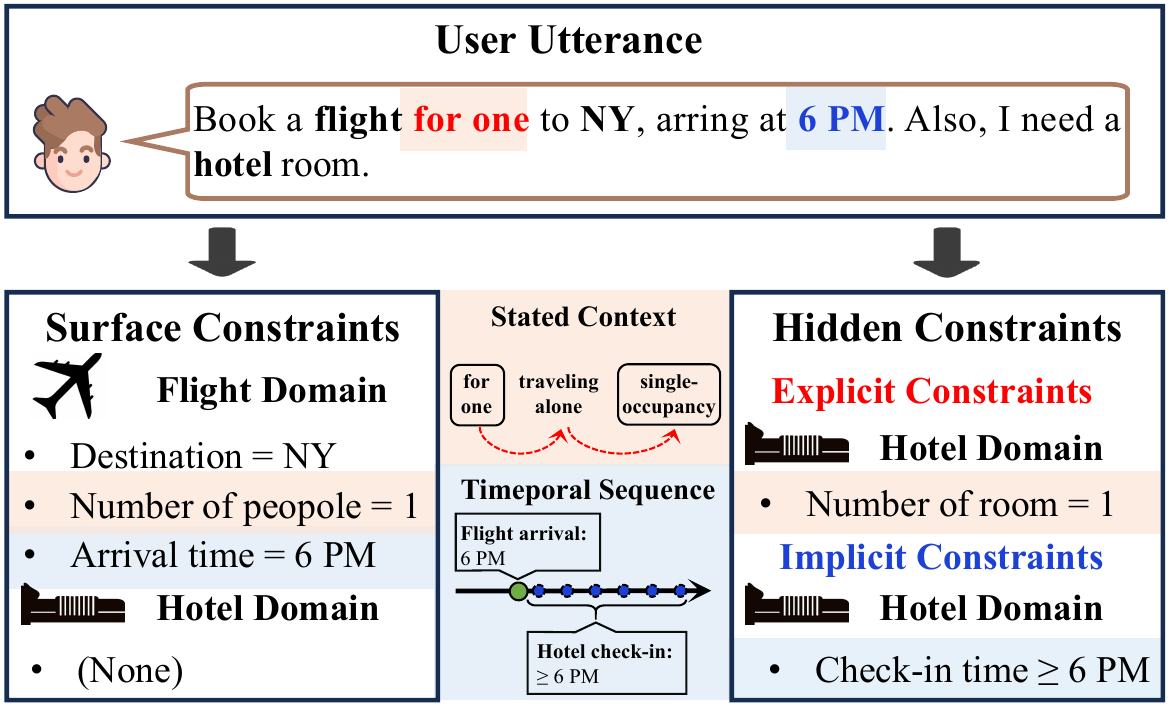}
	\caption{Example of \emph{explicit} (traveling alone implies single-occupancy accommodation) and \emph{implicit} (hotel check-in must follow flight arrival) constraints in a cross-domain scenario.}
	\label{example_intro}
\end{figure}

Task-oriented dialogue systems have shown strong performance in single-domain and loosely coupled multi-domain settings \cite{20}. However, cross-domain dialogues, where decisions in one domain impose feasibility constraints on others, remain particularly challenging \cite{ZhuHZZH20, 0001PCLY0CL23}. These constraints are often only partially stated and may be \emph{explicit} or \emph{implicit}, requiring commonsense or temporal reasoning, as illustrated in Fig.~\ref{example_intro}. Though rarely expressed verbatim, these constraints are critical for ensuring globally valid actions across tasks.

Existing methods largely address this problem through either dialogue state construction or policy optimization. Dialogue state tracking aims to uncover latent user information within a predefined ontology \cite{DongFLS024, LinLMMZCWYCSF21}, but they are brittle when cross-domain constraints are not directly grounded in surface text or require commonsense inference. On the other hand, multi-task or hierarchical reinforcement learning (RL) improves long-horizon decision making \cite{27, 28}, yet typically assumes accurate and complete states and degrades when critical constraints are missing \cite{ZhaoDLWW24}. Together, these observations suggest a bottleneck: cross-domain policy learning is fundamentally constrained by whether the dialogue state captures (implicit and explicit) feasibility constraints.

\begin{figure*}[t]
	\centering
	\includegraphics[width=2\columnwidth]{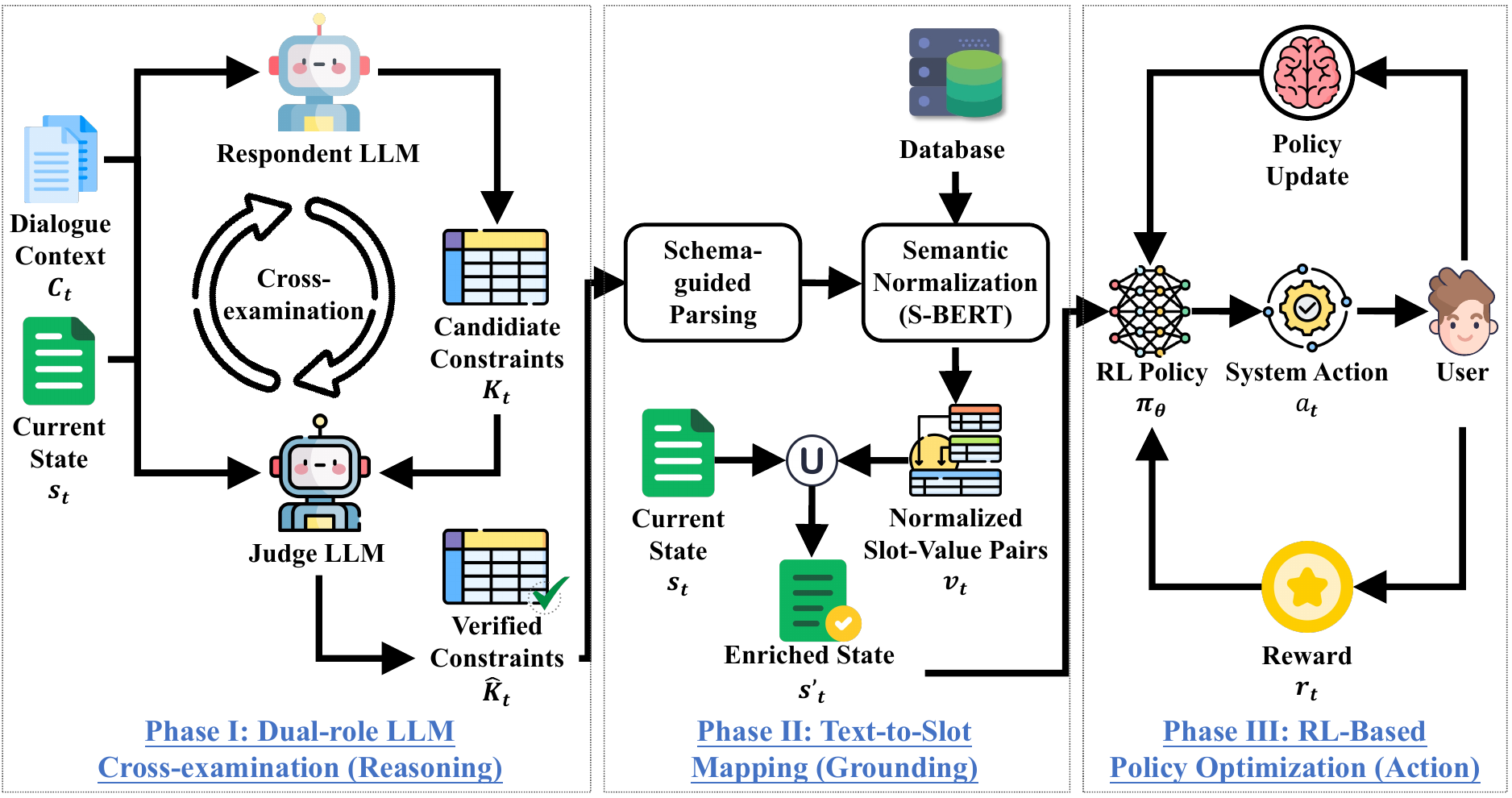}
	\caption{Overview of the proposed VLK-RL framework. Candidate constraints are inferred from the respondent LLM and verified by dual-role cross-examination, then grounded into structured normalized slot-value pairs to enrich the dialogue state for downstream RL policy optimization.}
	\label{fig:framework}
\end{figure*}

Large language models (LLMs) offer a promising source of such constraint knowledge, as they encode substantial commonsense knowledge and can infer such constraints from context \cite{12, 17}. Nevertheless, using LLMs as end-to-end dialogue agents or directly integrating them into long-horizon RL policy pipelines is unreliable for task-oriented systems \cite{abs-2402-18013, nguyen2025spec}. LLM outputs may hallucinate and drift across turns, and their free-form generations are hard to verify, align with ontologies, or safely encode as state inputs. Consequently, naively coupling LLM reasoning with RL policy can corrupt state representations and misguide policy learning over long horizons.

We address this gap by reframing constraint modeling as \emph{constraint-aware dialogue state construction} and propose \textbf{V}erified \textbf{L}LM-\textbf{K}nowledge empowered \textbf{RL} (\textbf{VLK-RL}), a hybrid framework that makes LLM-derived constraint reasoning reliable and actionable for RL policies. Given a dialogue context, VLK-RL first elicits candidate explicit and implicit feasibility constraints with an LLM. A key challenge is that such inferences are not directly trustworthy for long-horizon decision making. To this end, we introduce a \emph{dual-role cross-examination} mechanism inspired by fact verification \cite{35}, in which two LLMs assume differentiated roles—a \textit{respondent} and a \textit{judge}—to collaboratively validate inferred constraints through cross-examination dialogue without external supervision. Even after verification, constraints remain free-form and cannot be directly consumed by downstream RL. We therefore design a \emph{text-to-slot mapper} that grounds verified constraints into structured normalized representations compatible with downstream RL. The resulting structured, constraint-aware states can be consumed by standard RL policies without changing policy architectures, enabling robust long-horizon planning under cross-domain feasibility. We evaluate VLK-RL on MultiWOZ~2.1 and Frames, and observe consistent gains in cross-domain generalization and policy robustness over strong cross-domain single-model baselines.

In summary, our contributions are threefold:
\vspace{-3pt}
\begin{itemize}
\item We identify feasibility-constraint completeness as a central bottleneck for cross-domain task-oriented dialogue, and formulate constraint modeling as a state construction problem that unifies explicit and implicit dependencies across domains.
\vspace{-3pt}
\item We propose VLK-RL, which integrates LLM reasoning with RL policy optimization, while ensuring both inferred knowledge reliability and modality alignment without external supervision.
\vspace{-3pt}
\item Extensive experiments on multiple benchmarks demonstrate that injecting verified, structured constraint knowledge improves cross-domain generalization  and robustness.
\end{itemize}

\section{Method}

As shown in Fig.~\ref{fig:framework}, VLK-RL decouples \emph{reasoning} from \emph{control} via a modular state-construction interface with three components:
(1) a \textit{dual-role LLM cross-examination module} that verifies explicit and implicit constraints via cross-examination dialogue;
(2) a \textit{text-to-slot mapper} that grounds verified constraints into structured normalized slot-value pairs and augments the dialogue state from $s_t$ to an enriched state $s'_t$; and
(3) an \textit{RL-based policy} that makes long-horizon decisions conditioned on $s'_t$.

\subsection{Dual-role LLM Cross-examination Module}

Although LLMs possess world knowledge and strong commonsense reasoning ability, their outputs often contain hallucinations and inconsistencies. Directly injecting such unreliable reasoning into policy learning risks severe performance degradation. To address this, we introduce a \textit{dual-role cross-examination module}, inspired by \citet{35}, where two LLMs assume complementary roles—\textit{respondent} and \textit{judge}—and engage in interactive reasoning to filter factual errors. The core rationale of this module is that interactions between different roles can reveal hidden inconsistencies in reasoning. Such inconsistencies can be considered as signals of uncertainty in the respondent’s original claims, and thus provide a critical basis for judging whether its inferences are correct.

\begin{figure}[t]
	\centering
	\includegraphics[width=1\columnwidth]{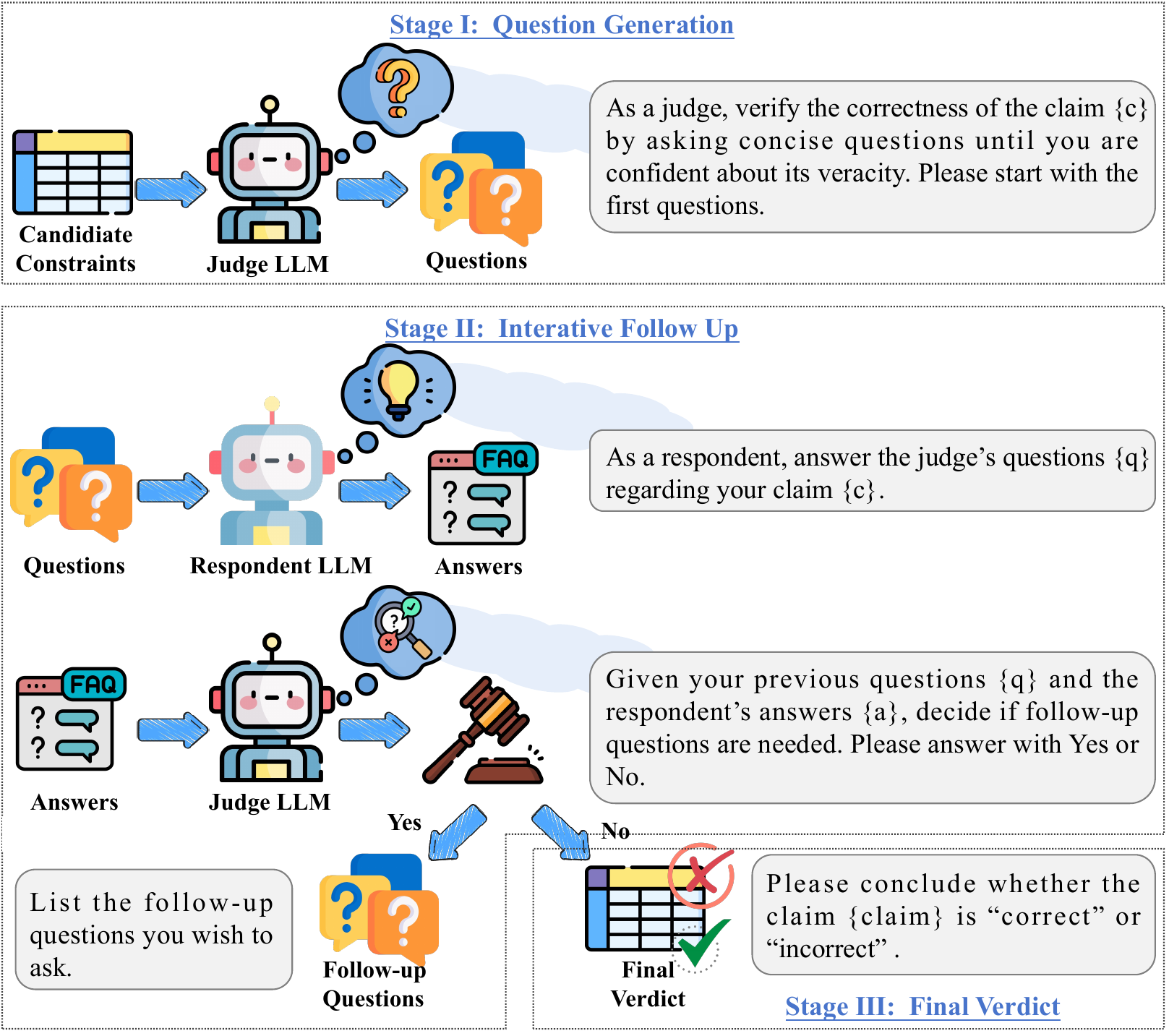}
	\caption{Dual-role cross-examination. A \textit{respondent} proposes constraints; a \textit{judge} probes with targeted questions and returns a verification verdict, filtering hallucinations and cross-turn inconsistencies.}
	\label{dualLLM}
\end{figure}

\paragraph{Respondent.}
The respondent LLM takes the dialogue context $C_t$ and current state $s_t$ as input and proposes a set of candidate constraint inferences $K_t = \mathcal{R}(C_t, s_t)$. It covers both explicit constraints and implicit constraints.
We design structured CoT prompts (see App.~\ref{app:A} for details) to guide the respondent LLM in identifying domains, extracting facts, and deriving candidate constraints, thereby improving both reasoning accuracy and interpretability by exposing intermediate steps. During cross-examination, the respondent LLM further defends its claims by answering the judge’s challenges to simulate dialogue-based validation.

\paragraph{Judge.}
Conditioned on $(C_t, K_t)$, the judge LLM evaluates each candidate $k \in K_t$ through an interactive probing procedure. Concretely, the cross-examination proceeds in three prompted stages, as shown in Fig.~\ref{dualLLM}:

In Stage I (\textit{question generation}), the judge LLM formulates clarification questions regarding the claims of respondent LLM, which are sequentially answered by the respondent and appended to the context. 
In Stage II (\textit{iterative follow-up}), the judge LLM inspects the answers and, if contradictions or gaps are detected, issues further questions until no new queries arise or a predefined round limit is reached (set to 5 in our experiments). 
In Stage III (\textit{final verdict}), the judge outputs a binary decision (\texttt{True}/\texttt{False}) for each constraint claim, indicating whether it is logically valid or unsupported. 
Intuitively, contradictions or evasive responses in this process are treated as signals of uncertainty, leading to the rejection of the corresponding inference. Only the subset of verified constraints endorsed by the judge LLM, denoted $\hat{K}_t$, are retained as reliable knowledge for downstream processing.

\subsection{Text-to-Slot Mapper}

Although the verified constraints $\hat{K}_t$ are reliable, they remain expressed in natural language, which is incompatible with structured policy representations. Moreover, the raw outputs of LLMs may contain values that do not exactly correspond to entries in the TOD database (e.g., ‘NYC downtown' versus the ontology term ‘Manhattan'), which would cause execution failures.

To bridge this gap, we design a text-to-slot mapper $\mathcal{M}: \hat{K}_t \to V_t$, where $V_t = \{ (s_i, v_i) \}$ denotes a set of slot–value pairs consistent with the TOD database. Each verified constraint $k \in \hat{K}_t$ is first parsed into candidate slot–value pairs using schema-guided extraction rules and ontology-based templates. This rule-driven approach requires no additional training and ensures that free-form natural language descriptions are aligned with the predefined slot schema across domains.

To resolve the mismatch between LLM-generated values and database-predefined values, we perform semantic similarity-based normalization, guaranteeing that all slot–value pairs are fully compatible with the database entries. For each $(s, v)$, if $s \in DB$, we retrieve valid values for $s$ and compute embedding similarity with $v$ using Sentence-BERT \cite{36}. 
The most similar entry is selected:

\[
\tilde{v} = \arg\max_{v' \in DB(s)} \cos(\mathbf{e}(v), \mathbf{e}(v')).
\]
If $\cos(\mathbf{e}(v), \mathbf{e}(\tilde{v})) \geq \tau$ (with $\tau = 0.7$), $v$ is replaced with $\tilde{v}$; otherwise, the pair is discarded. 
If $s$ is not in the ontology, we first match the most similar slot and then normalize its value. 
Finally, normalized slot–value pairs (e.g., \texttt{hotel\_area = Midtown Manhattan}) are integrated into the dialogue state:
\[
s'_t = s_t \cup v_t,
\]
where $s_t$ is the original state and $v_t$ contains structured, verified knowledge. 
The enriched state $s'_t$ is then passed to the RL-based policy optimizer for robust decision-making.

\subsection{RL-based Policy Optimizer}

Although LLMs excel at knowledge reasoning, they struggle with long-horizon decision-making. RL, on the other hand, provides a principled framework for optimizing policies based on long-horizon rewards. By combining RL with validated knowledge from LLMs, our framework aims to achieve both policy robustness and cross-domain generalization.

We formulate TOD policy optimization as a Markov Decision Process defined by $(\mathcal{S}, \mathcal{A}, P, R, \gamma)$. 
The state $s \in \mathcal{S}$ is the enriched dialogue state, which encodes the dialogue context, historical slot–value pairs, validated cross-domain constraints, and dialogue turn information. 
The action $a \in \mathcal{A}$ corresponds to a system response action. 
The reward $R(s,a)$ follows the MultiWOZ convention: completing all domain goals yields a reward of $2L$, completing only a single domain yields $+5$, failure yields $-L$, and each intermediate turn incurs a $-1$ penalty to encourage concise dialogues. 
To update the policy, we adopt Proximal Policy Optimization (PPO) as a representative RL algorithm due to its stability and strong empirical performance in dialogue policy learning. The clipped surrogate objective is given by:

\begin{multline}
	\label{PPO}
	L^{\mathrm{CLIP}}(\theta) 
	= \hat{\mathbb{E}}_t \Big[ 
	\min\big( r_t(\theta)\,\hat{A}_t^{s'}, \\ 
	\mathrm{clip}(r_t(\theta),1-\epsilon,1+\epsilon)\,\hat{A}_t^{s'} \big)
	\Big]
\end{multline}

\noindent where$r_t(\theta) \;=\; \frac{\pi_{\theta}(a_t \mid s'_t)}{\pi_{\theta_{\text{old}}}(a_t \mid s'_t)}$ is the probability ratio between the updated and the previous policy on the enriched state \(s'_t\). \(\hat{A}_t^{s'}\) denotes the advantage estimate for the enriched state \(s'_t\) (i.e., the state that incorporates both historical slots and the verified, normalized slot--value pairs). The expectation \(\hat{\mathbb{E}}_t\) is taken over timesteps in a sampled minibatch, and \(\epsilon\) is the clipping hyperparameter.

VLK-RL enhances $s_t$ with verified feasibility constraints by modifying only state construction, while keeping the policy architecture, action space, and optimization objective unchanged. As a result, it is fully compatible with standard RL policy learning. For concreteness, we instantiate the framework with PPO as our default optimizer. Importantly, other RL backbones can be plugged in without altering our verification or grounding modules, a claim we further validate in App.~\ref{appendix:rl-backbone}.

Alg.~\ref{alg:vlkrl} summarizes the workflow. At each turn, the framework first performs cross-examination to infer and validate constraints (lines 2–3), then maps the validated constraints into ontology-aligned slot–value pairs (line 4), merges them into the dialogue state (line 5), and finally selects the next system action through RL policy (lines 6–7).

\begin{algorithm}[t]
	\caption{VLK-RL Framework for Multi-domain TOD}
	\label{alg:vlkrl}
	\begin{algorithmic}[1]
		\REQUIRE Dialogue context $C$, database $DB$, RL policy $\pi$, dialogue state $s$
		\FOR{each dialogue turn $t$}
		\STATE $K_t \leftarrow$ Respondent LLM inference on $C_t$
		\STATE $\hat{K}_t \leftarrow$ Judge LLM verifies $K_t$
		\STATE $v_t \leftarrow \mathcal{M}(\hat{K}_t, DB)$ \COMMENT{text-to-slot mapping and normalization}
		\STATE $s'_t \leftarrow s_t \cup v_t$
		\STATE Sample action $a_t \sim \pi_\theta(\cdot \mid s'_t)$
		\STATE Execute system response action $a_t$
		\ENDFOR
	\end{algorithmic}
\end{algorithm}

\section{Experiments}

We evaluate VLK-RL on cross-domain task-oriented dialogue along three axes: 
(i) overall performance and robustness against strong baselines in simulated and human evaluations (Sec.~\ref{main} and Sec.~\ref{human}), 
(ii) the contribution of each module via ablations (Sec.~\ref{ablation}), and 
(iii) constraint-oriented analyses that combine explicit/implicit failure statistics with qualitative case studies (Sec.~\ref{sec:constraint_analysis}).
Additional studies (e.g., low-resource training (App.~\ref{app:scratch}), cross-model cross-examination (App.~\ref{app:cross_model}), and alternative RL backbones (App.~\ref{appendix:rl-backbone}) are deferred to the appendix \footnote{Code and data are publicly available at: \url{https://github.com/amarantosQWQ/VLK-RL. }}.

\subsection{Datasets}
We conduct experiments on two widely used multi-domain benchmarks: MultiWOZ~2.1 \cite{33} and Frames\footnote{\url{https://datasets.maluuba.com/Frames.}}. 
MultiWOZ~2.1 contains over 10k human--human dialogues spanning seven domains with annotated slot--value states, and exhibits rich cross-domain dependencies.
Frames is a Wizard-of-Oz dataset with interrelated subtasks and strong cross-task feasibility requirements; following \citet{PengLLGCLW17}, we adopt a modified schema that encodes inter-subtask constraints and preferences.
We use ConvLab-2\footnote{\url{https://github.com/thu-coai/ConvLab-2.}} for simulation, database access, and evaluation to ensure reproducibility and fair comparison with prior work\footnote{We use ConvLab-2 for stable RL training; ConvLab-3 RL pipelines remain less stable in our setting (see repository issue  \#179 and \#191 discussions).}.

\begin{table*}[h]
	\centering
	\caption{Performance Comparison of different dialogue agents on MultiWOZ 2.1 and Frames. Top performance per metric is highlighted with bold and a background. All differences statistically significant ($p<0.05$).}
	\label{main_results}
	\scalebox{0.66}{
		\begin{tabular}{l l c c c c c c c}
			\hline
			Dataset & Model & Avg. Precision & Avg. F1 & Avg. Recall & Complete/Tot & Success/Tot & Avg. Turn (Succ) & Avg. Turn (All) \\
			\hline
			\multirow{9}{*}{\centering MultiWOZ 2.1} & PPO             & 0.4273 & 0.4997 & 0.7121 & 0.4912 & 0.3815 & 13.21 & 20.94 \\	
			& ACGOS           & 0.4857 & 0.5328 & 0.7316 & 0.5524 & 0.4521 & 14.10 & 19.82 \\
			& GALAXY          & 0.5236 & 0.5789 & 0.7654 & 0.6031 & 0.5216 & 13.50 & 21.56 \\
			& GDP-Zero        & {0.5819} & {0.6357} & 0.7928 & {0.7025} & {0.6024} & 15.31 & 22.22 \\
			& TransferTOD     & 0.5648 & 0.6105 & 0.7817 & 0.6716 & 0.5823 & 14.83 & 20.40 \\
			& CAPID           & 0.5763 & 0.6152 & {0.7871} & 0.6820 & 0.5875 & 14.00 & 20.00 \\
			& VLK-RL (GPT-4o-mini) & 0.6354 & 0.6886 & 0.8237 & 0.7619 & 0.6812 & 13.00 & 17.91 \\
			& VLK-RL (Qwen-7B)      & 0.6483 & 0.7027 & 0.8354 & 0.7815 & 0.6958 & 12.80 & 17.62 \\ 
			& \cellcolor{blue!5}\textbf{VLK-RL (Qwen-14B)} & \cellcolor{blue!5}\textbf{0.6628} & \cellcolor{blue!5}\textbf{0.7182} & \cellcolor{blue!5}\textbf{0.8429} & \cellcolor{blue!5}\textbf{0.8006} & \cellcolor{blue!5}\textbf{0.7214} & \cellcolor{blue!5}\textbf{12.51} & \cellcolor{blue!5}\textbf{17.35} \\
			\hline
			\multirow{9}{*}{\centering Frames}  & PPO             & 0.4231 & 0.4802 & 0.6721 & 0.6031 & 0.4235 & 15.34 & 18.56 \\
			& ACGOS           & 0.4852 & 0.5331 & 0.7309 & 0.5236 & 0.4315 & 13.50 & 18.80 \\
			& GALAXY          & 0.5186 & 0.5734 & 0.7598 & 0.5810 & 0.4975 & 12.90 & 20.10 \\
			& GDP-Zero        & {0.5819} & {0.6327} & 0.7578 & {0.7215} & {0.5890} & 14.12 & 17.84 \\
			& TransferTOD     & 0.5596 & 0.6087 & 0.7803 & 0.6507 & 0.5678 & 13.80 & 19.60 \\
			& CAPID           & 0.5712 & 0.6154 & {0.7883} & 0.6701 & 0.5782 & 14.00 & 20.00 \\
			& VLK-RL (GPT-4o-mini) & 0.6302 & 0.6875 & 0.8204 & 0.7512 & 0.6789 & 12.30 & 17.80 \\
			& VLK-RL (Qwen-7B)      & 0.6437 & 0.6989 & 0.8325 & 0.7701 & 0.6903 & 12.10 & 17.50 \\ 
			& \cellcolor{blue!5}\textbf{VLK-RL (Qwen-14B)} & \cellcolor{blue!5}\textbf{0.7034} & \cellcolor{blue!5}\textbf{0.7512} & \cellcolor{blue!5}\textbf{0.8357} & \cellcolor{blue!5}\textbf{0.8063} & \cellcolor{blue!5}\textbf{0.7239} & \cellcolor{blue!5}\textbf{12.65} & \cellcolor{blue!5}\textbf{15.91} \\
			\hline
		\end{tabular}
	}
\end{table*}

\subsection{Experimental Setup}
Unless stated otherwise, we follow ConvLab-2 defaults. Key hyperparameters are: training epochs $=300$, maximum dialogue length $L=30$, batch size $=100$, cross-examination rounds $R=5$, and normalization threshold $\tau=0.7$. We use PPO as the RL optimizer. For the LLM backbone, we evaluate Qwen2-7B-Instruct, Qwen1.5-14B-Chat (GPTQ-Int4), and GPT-4o-mini to cover different capability and deployment regimes. The Qwen family provides strong reasoning for multi-turn dialogue and supports efficient local deployment via quantization, while GPT-4o-mini represents a high-performance commercial model, serving as a strong reference point. By default, the \textit{judge} and \textit{respondent} roles share the same backbone with role-specific prompts, and we report cross-model variants in App.~\ref{app:cross_model}. 
All LLMs are used off-the-shelf without task-specific fine-tuning. All results are averaged over 5 runs with different random seeds.

\subsection{Evaluation Metrics}
We report standard ConvLab-2 metrics: average dialogue-act Precision/Recall/F1, task Complete and Success rates, and average turns for successful dialogues and for all dialogues (lower is better). 
Note that shorter dialogues may reflect premature failures rather than efficiency, hence these metrics are interpreted jointly.  
Following common practice, we interpret success/complete jointly with turns to reflect both task achievement and efficiency.

\subsection{Baselines}
We compare VLK-RL with RL-based cross-domain policy baselines (PPO \cite{34}, ACGOS \cite{7}), LLM-based dialogue models (GALAXY \cite{17}, GDP-Zero \cite{18}, TransferTOD \cite{12}), and a cross-domain DST baseline (CAPID \cite{DongFLS024}).

\subsection{Main Results}

\subsubsection{Simulated Environments}
\label{main}

Tab.~\ref{main_results} reports results on MultiWOZ~2.1 and Frames. Across both datasets, VLK-RL achieves the strongest overall performance, with consistent improvements in Complete and Success and fewer average turns, indicating that it completes more user goals and reduces redundant interactions. The improvement is notably larger on Frames, where subtasks are more tightly coupled and feasibility constraints often determine whether a plan is globally valid, supporting our claim that explicitly modeling and enforcing cross-domain constraints is central to robust long-horizon decision making. RL-only baselines (PPO, ACGOS) remain limited, especially on Frames, suggesting that long-horizon exploration and credit assignment alone are insufficient to recover missing feasibility knowledge from raw interaction. LLM-based approaches (GALAXY, GDP-Zero, TransferTOD) achieve stronger dialogue-act metrics, but their end-to-end success lags behind VLK-RL, consistent with the brittleness of unverified or weakly grounded single-pass reasoning that can accumulate errors across turns. CAPID improves cross-domain state tracking and narrows the gap, yet it primarily strengthens states with information grounded in dialogue and ontology, so implicit feasibility constraints that require commonsense or temporal inference can still be under-specified, limiting downstream policy success. In contrast, VLK-RL explicitly verifies candidate constraints and grounds them into ontology-aligned slot--value representations that are executable against the database, yielding a more stable state interface for policy learning and mitigating drift over long dialogues. Among VLK-RL variants, stronger backbones tend to perform better, and we observe that Qwen-based variants achieve the highest scores in our setting; we attribute this mainly to more consistent constraint-to-slot grounding and fewer cross-turn inconsistencies under our prompts, and we provide additional analysis and cross-model cross-examination results in App.~\ref{app:cross_model}.

\begin{figure*}[t]
	\centering
	\includegraphics[width=2.1\columnwidth]{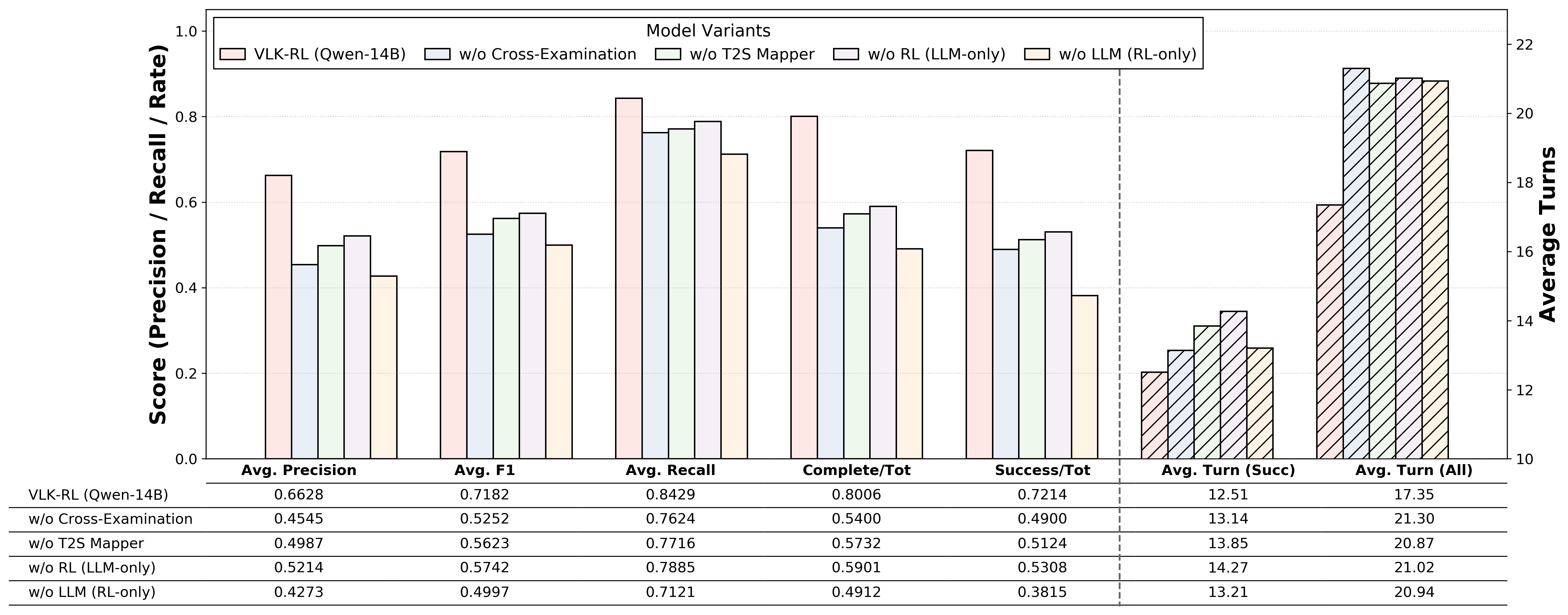}
	\caption{Ablation study on VLK-RL (Qwen-14B) to analyze the individual contributions of its three core modules.}
	\label{fig:ablation}
\end{figure*}

\subsubsection{Human Environments}
\label{human}

To assess robustness in realistic interactions, we conduct human evaluation with 30 annotators on MultiWOZ~2.1 and Frames. For each dataset, we use ConvLab-2's \texttt{goal\_generator} to sample user goals, and each annotator completes three dialogues by interacting with anonymized agents whose identities are shuffled and hidden to avoid bias. After each dialogue, the annotator provides two judgments: task completion as \textsc{Success Rate} (SR), and overall dialogue quality as \textsc{Human Rating} (HR) on a 1--5 Likert scale (higher is better), considering fluency, naturalness, and redundancy. Annotators are allowed to terminate dialogues early when interactions become incoherent or unproductive; such cases are counted as failures for SR, and the corresponding HR is still recorded to reflect perceived quality under failure. 

Tab.~\ref{tab:human_eval} shows that VLK-RL consistently achieves higher SR and HR than all baselines on both datasets, with larger gains on Frames where cross-task feasibility constraints are more salient. Compared with RL-only agents, VLK-RL reduces redundant clarification turns and improves cross-domain coordination; compared with LLM-based and DST-based baselines, verifying and grounding constraints at the state level helps maintain feasibility over long horizons, improving both task validity and human-perceived coherence. These human results corroborate the simulated evaluation and support the central claim that constraint verification and ontology-aligned grounding provide a reliable interface for long-horizon policy learning.

\begin{table}[h]
	\centering
	\caption{Human evaluation results of different agents.}
	\label{tab:human_eval}
	\scalebox{0.7}{
		\begin{tabular}{l l c c}
			\hline
			\textbf{Dataset} & \textbf{Model} & \textbf{SR} & \textbf{HR} \\
			\hline
			\multirow{9}{*}{\centering MultiWOZ 2.1} & PPO                  & 0.2850 & 2.21 \\
			& ACGOS                & 0.3410 & 2.35 \\
			& GALAXY               & 0.3875 & 2.46 \\
			& GDP-Zero             & 0.4120 & 2.61 \\
			& TransferTOD          & 0.3982 & 2.58 \\
			& CAPID                & 0.4056 & 2.63 \\
			& VLK-RL (GPT-4o-mini) & 0.4713 & 3.04 \\
			& VLK-RL (Qwen-7B)     & 0.4936 & 3.07 \\
			& \cellcolor{blue!5}\textbf{VLK-RL (Qwen-14B)} & \cellcolor{blue!5}\textbf{0.5124} & \cellcolor{blue!5}\textbf{3.18} \\
			\hline
			\multirow{9}{*}{\centering Frames} & PPO                  & 0.2618 & 2.08 \\
			& ACGOS                & 0.3185 & 2.22 \\
			& GALAXY               & 0.3612 & 2.34 \\
			& GDP-Zero             & 0.3896 & 2.49 \\
			& TransferTOD          & 0.3741 & 2.45 \\
			& CAPID                & 0.3927 & 2.51 \\
			& VLK-RL (GPT-4o-mini) & 0.4589 & 3.11 \\
			& VLK-RL (Qwen-7B)     & 0.4823 & 3.16 \\
			& \cellcolor{blue!5}\textbf{VLK-RL (Qwen-14B)} &\cellcolor{blue!5} \textbf{0.5057} & \cellcolor{blue!5}\textbf{3.32} \\
			\hline
		\end{tabular}
	}
\end{table}

\subsection{Ablation Study}
\label{ablation}

We analyze the contribution of each component in VLK-RL using the best-performing variant, VLK-RL (Qwen-14B), and report results in Fig.~\ref{fig:ablation}. We consider four ablations. (1) \textbf{w/o Cross-Examination} removes the dual-role verification and uses the respondent LLM outputs directly as constraints. (2) \textbf{w/o T2S Mapper} removes ontology-aligned grounding and instead encodes verified textual constraints into dense representations that are concatenated with the dialogue state, without slot normalization. (3) \textbf{w/o RL (LLM-only)} removes RL optimization and prompts the LLM to select an action from the predefined action set (prompt in App.~\ref{app:B}). (4) \textbf{w/o LLM (RL-only)} removes both LLM-based modules, reducing the system to PPO on the original state.

\begin{figure}[t]
	\centering
	\includegraphics[width=1\columnwidth]{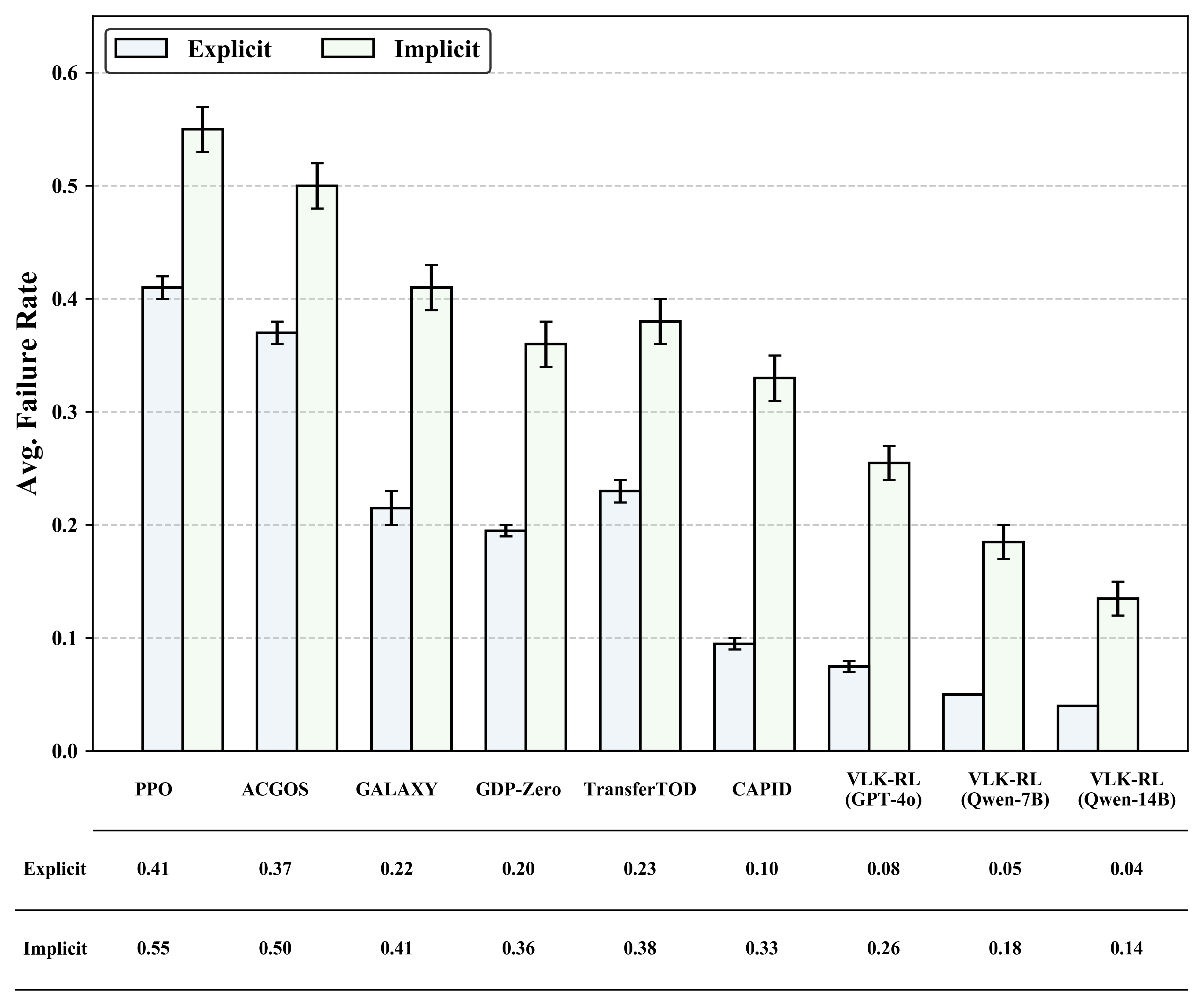}
	\caption{Constraint-related failure breakdown on MultiWOZ~2.1 and Frames. For failed dialogues, we report the fraction whose failure is attributable to missing \emph{explicit} constraints or missing \emph{implicit} constraints. Per-dataset breakdowns are provided in App.~\ref{app:static}.}
	\label{static:si}
\end{figure}

Removing any component substantially degrades performance, indicating that VLK-RL’s gains arise from the interaction between verification, state grounding, and long-horizon optimization rather than from any single dominant module. In particular, \textbf{w/o Cross-Examination} sharply reduces success and precision, showing that hallucinations and cross-turn inconsistencies in single-pass constraint inference corrupt the state signal used for policy learning, motivating explicit verification. \textbf{w/o T2S Mapper} also leads to a pronounced drop, despite using verified constraints, because free-form text fails to provide an executable and stable interface to the ontology and database; without slot-level grounding and value normalization, constraints may be non-fillable, schema-inconsistent, or ambiguous across domains, weakening their utility for downstream RL. The \textbf{LLM-only} variant achieves moderate act-level quality but lags in end-to-end task success and turns, reflecting that myopic action selection without RL struggles to maintain goal consistency and recover from earlier errors over long horizons. Finally, the \textbf{RL-only} variant performs worst overall, confirming that long-horizon optimization alone cannot compensate for missing implicit feasibility constraints in the state.

To make these failure modes concrete, App.~\ref{app:t2s_mapper} quantifies grounding errors and shows that a substantial fraction of verified constraints cannot be executed reliably against the ontology and database without slot-value mapping and normalization, explaining the drop for w/o T2S Mapper. App.~\ref{app:cross_exam} analyzes verification errors and reports that cross-examination filters hallucinated constraints, leading to higher-precision state augmentation and more stable long-horizon policy learning.

\vspace{-5pt}
\subsection{Constraint-oriented Analyses}
\label{sec:constraint_analysis}

We further analyze whether agents satisfy cross-domain feasibility constraints by combining explicit and implicit failure statistics with qualitative inspection. Following prior TOD analyses, we distinguish \emph{explicit} constraints that are directly stated or trivially implied by user utterances from \emph{implicit} constraints that require commonsense, temporal, or inter-domain reasoning. Concretely, we sample 50 user goals from each dataset and run each system to obtain 250 dialogues per dataset. For each failed dialogue, we attribute the failure to missing explicit or implicit constraints and report, among failures, the fraction attributable to each type.

\begin{table*}[t]
	\centering
	\caption{Dialogue examples for a multi-domain travel planning task. Comparison between PPO and VLK-RL.  
		Red text highlights implicit associations (inferred from dialogue context, such as dates and locations), while green text highlights explicit associations (direct responses to user constraints).  
		(U = User, S = System).}	
	\label{tab:dialogue_examples2}
	\renewcommand{\arraystretch}{1.2}
	\scalebox{0.9}{
		\begin{tabular}{p{0.47\textwidth} p{0.47\textwidth}}
			\hline
			\textbf{PPO} & \textbf{VLK-RL} \\
			\hline
			U0: Hi, I need to go to Cambridge from London on Wednesday morning. I'd like to go by myself. \par
			S1: Sure, I can help you with that. There is a train leaving from London Liverpool Street at 09:15 on Wednesday, heading to Cambridge. Would you like to book a ticket? \par
			U2: No need to book, just give me the train information. Also, I’m planning to visit the Cambridge University Botanic Gardens. \par
			S3: Okay, so what's the area of the Cambridge University Botanic Gardens? \par
			U4: Well, the area is definitely Bateman Street, hasn't it? \par
			S5: Okay, the area is Bateman Street. \par
			U6: Oh, by the way, I also need to book a hotel. I'm looking for a four-star hotel, and it should be on the east side of the scenic area. \par
			S7: Okay, for the hotel booking, what are the dates and how many people will it be for? Also, do you have any parking requirements? \par
			U8: The hotel is for Wednesday, for one person, and I don't need parking. \par
			... &
			U0: Hi, I need to go to Cambridge from London on Wednesday morning. I'd like to go by myself. \par
			S1: Sure, I can help you with that. There is a train leaving from London Liverpool Street at 09:15 on Wednesday, heading to Cambridge. Would you like to book a ticket? \par
			U2: No need to book, just give me the train information. Also, I’m planning to visit the Cambridge University Botanic Gardens. \par
			S3: Alright, the area of the Cambridge University Botanic Gardens should be \textcolor{red}{Bateman Street} (implicit association). Do you need a hotel reservation for \textcolor{red}{Wednesday} (implicit association)? I can book \textcolor{green}{a single room for you} (explicit association). Also, will you require a parking space? \par
			U4: That would be great, please book a hotel for me. I won't need a parking space. Oh, by the way, it needs to be a four-star hotel on the east side of the scenic area. \par
			... \\
			\hline
		\end{tabular}
	}
\end{table*}

Fig.~\ref{static:si} summarizes the results. RL-only baselines exhibit high failure rates on both constraint types, reflecting their reliance on incomplete states and the difficulty of enforcing feasibility purely through long-horizon optimization. LLM-based baselines reduce explicit failures but remain unstable on implicit constraints, consistent with reasoning errors and cross-turn inconsistencies accumulating in long interactions. CAPID achieves relatively low explicit failure rates due to stronger ontology-grounded state tracking, yet implicit failures remain high because such constraints are rarely grounded in surface text and not explicitly represented at the state level. In contrast, VLK-RL substantially reduces failures for both explicit and implicit constraints across datasets, indicating that verifying and grounding LLM-inferred constraints into slot--value states improves the system’s ability to recognize and enforce feasibility throughout the dialogue trajectory. We observe the largest relative improvements on Frames, which contains tightly coupled subtasks and makes feasibility constraints more consequential to end-to-end success.

%To complement the aggregate statistics, we provide qualitative case studies in App.~\ref{app:D}. These examples illustrate how verified constraints are surfaced during cross-examination, how they are grounded into ontology-aligned slot--value updates, and how the resulting constraint-aware state leads to fewer redundant clarification turns and fewer globally invalid decisions, especially in cases where the user does not state inter-domain dependencies explicitly.

\subsection{Case Study}
\label{case_study}

To complement the quantitative results, we present a case study comparing PPO and VLK-RL in a multi-domain travel planning scenario.  
As shown in Tab.~\ref{tab:dialogue_examples2}, the task involves coordinating transportation, attraction visits, and hotel booking under realistic user constraints.

The PPO agent struggles with implicit associations: although it retrieves the train information, it fails to leverage contextual cues to infer the attraction’s location or connect the date with the hotel booking request. This results in redundant clarification questions (e.g., asking for the obvious location of Cambridge University Botanic Gardens) and longer dialogues. In contrast, VLK-RL successfully resolves both explicit and implicit constraints. It infers that the attraction is on Bateman Street and proactively associates the hotel booking with the same day (Wednesday), while explicitly confirming the user’s room type request. These mechanisms reduce unnecessary turns and improve dialogue coherence. Overall, the case study confirms that VLK-RL enhances policy efficiency by grounding decisions in both explicit slot-value matches and implicit contextual reasoning, leading to smoother and more natural multi-domain dialogues.

\section{Related Work}

Task-oriented dialogue systems in cross-domain settings require managing long-horizon dependencies, where decisions in one domain impose feasibility constraints on others. Traditional cross-domain dialogue state tracking (DST) methods, such as TRADE \cite{WuMHXSF19} and TripPy \cite{HeckNLGLMG20}, focus on extracting slot-value representations to model user goals. While effective for explicit constraints, these methods struggle to handle implicit constraints, which require commonsense or temporal reasoning (e.g., hotel check-in must follow flight arrival). Recent work has extended DST with large language models (LLMs) to improve generalization across domains, allowing LLMs to infer both explicit and implicit constraints from the dialogue context \cite{DongFLS024, FengLLZW23, abs-2304-04256}. However, LLM-generated outputs often lack grounding and verifiability, making them challenging to integrate with structured databases and decision-making pipelines. Another line of research addresses cross-domain decision-making through reinforcement learning (RL). Composite-task and hierarchical RL frameworks aim to decompose complex goals into smaller sub-tasks and improve decision-making across multiple domains \cite{PengLLGCLW17, 24, 25}. These approaches typically assume that the dialogue state is accurate and complete. However, when cross-domain constraints are missing or misrepresented, RL policies become brittle, especially in long-horizon tasks where decisions in one domain influence others.

Recent hybrid LLM-RL frameworks have combined the strengths of LLMs and RL to bridge high-level reasoning and sequential decision-making. These methods use LLMs for planning, reward assignment, or task scheduling, while RL is responsible for decision-making \cite{turn1search7, turn1academia20, CaoZCSCLLZYL25}. However, many of these approaches tightly couple reasoning and action selection, which can lead to instability, especially in long-horizon tasks. In contrast, our approach decouples reasoning from control by using LLMs for constraint extraction, which are grounded into normalized states for RL optimization. This modular design allows LLM reasoning to inform RL policy without entangling the two processes, enhancing robustness and stability in cross-domain decision-making.

\section{Conclusion}

We present VLK-RL, a hybrid framework for cross-domain task-oriented dialogue that connects LLM reasoning and RL decision making through verified, state-level constraint grounding. VLK-RL verifies LLM-inferred explicit and implicit feasibility constraints via dual-role cross-examination and grounds the verified knowledge into ontology-aligned slot--value states, providing a stable and executable interface for downstream policy optimization. Experiments on MultiWOZ~2.1 and Frames show that VLK-RL consistently improves cross-domain generalization, policy robustness, and dialogue efficiency over strong RL-only, LLM-only, and cross-domain DST baselines in both simulated and human evaluations. Overall, VLK-RL offers a modular and principled approach to integrating verified LLM knowledge into long-horizon dialogue policy learning.

\section*{Limitation}

VLK-RL has several limitations. First, it relies on pre-trained LLMs for constraint inference and verification, which introduces additional latency and computation and may limit deployment in resource-constrained settings. Second, cross-examination improves reliability but is not guaranteed to be complete, and it may fail to surface subtle or rare commonsense dependencies, especially in complex contexts with ambiguous user goals. Third, the text-to-slot grounding depends on the coverage and granularity of the underlying ontology and database; constraints that cannot be cleanly expressed in the predefined schema may be partially lost. Future work may mitigate these issues by distilling the reasoning and verification modules into smaller models, incorporating retrieval or external knowledge to support rare constraints, and extending grounding mechanisms to handle richer constraint forms beyond slot--value representations.

\section*{Acknowledgments}
We are grateful to the anonymous reviewers for their insightful comments and valuable suggestions. We also sincerely appreciate the efforts of the human evaluators for their contributions to the manual assessment of our models.
This research was supported by the National Natural Science Foundation of China (Grant Nos. 62506046, 92570120, and 62306342) and the Hunan Provincial Natural Science Foundation (Grant No. 2024JJ6062). Additional support was provided by the Excellent Young Scientists Fund in Hunan Province (Grant No. 2024JJ4070), the Science and Technology Innovation Program of Hunan Province (Grant No. 2024RC3024), the Scientific Research Fund of Hunan Provincial Education Department (Grant No. 24B0001), and the Open Project of the Text Computing and Cognitive Intelligence Ministry of Education Engineering Research Center (Grant No. TCCI250101).

\bibliography{custom}

\begin{thebibliography}{28}
\providecommand{\natexlab}[1]{#1}

\bibitem[{Alon and David(2025)}]{turn1search7}
Yoav Alon and Cristina David. 2025.
\newblock \href
  {https://research-information.bris.ac.uk/en/publications/integrating-large-language-models-and-reinforcement-learning-for-}
  {Integrating large language models and reinforcement learning for non-linear
  reasoning}.
\newblock In \emph{FSE 2025}.

\bibitem[{Cao et~al.(2025)Cao, Zhao, Cheng, Shu, Chen, Liu, Liang, Zhao, Yan,
  and Li}]{CaoZCSCLLZYL25}
Yuji Cao, Huan Zhao, Yuheng Cheng, Ting Shu, Yue Chen, Guolong Liu, Gaoqi
  Liang, Junhua Zhao, Jinyue Yan, and Yun Li. 2025.
\newblock Survey on large language model-enhanced reinforcement learning:
  Concept, taxonomy, and methods.
\newblock \emph{{IEEE} Trans. Neural Networks Learn. Syst.}, 36(6):9737--9757.

\bibitem[{Cohen et~al.(2023)Cohen, Hamri, Geva, and Globerson}]{35}
Roi Cohen, May Hamri, Mor Geva, and Amir Globerson. 2023.
\newblock {LM} vs {LM:} detecting factual errors via cross examination.
\newblock In \emph{Proceedings of the 2023 Conference on Empirical Methods in
  Natural Language Processing, {EMNLP} 2023, Singapore, December 6-10, 2023},
  pages 12621--12640. Association for Computational Linguistics.

\bibitem[{Cordier et~al.(2022)Cordier, Urvoy, Lef{\`e}vre, and
  Rojas-Barahona}]{7}
Thibault Cordier, Tanguy Urvoy, Fabrice Lef{\`e}vre, and Lina~M Rojas-Barahona.
  2022.
\newblock Graph neural network policies and imitation learning for multi-domain
  task-oriented dialogues.
\newblock \emph{arXiv preprint arXiv:2210.05252}.

\bibitem[{Dong et~al.(2024)Dong, Feng, Lu, Shi, and Wu}]{DongFLS024}
Xiaoyu Dong, Yujie Feng, Zexin Lu, Guangyuan Shi, and Xiao{-}Ming Wu. 2024.
\newblock Zero-shot cross-domain dialogue state tracking via context-aware
  auto-prompting and instruction-following contrastive decoding.
\newblock In \emph{Proceedings of the 2024 Conference on Empirical Methods in
  Natural Language Processing, {EMNLP} 2024, Miami, FL, USA, November 12-16,
  2024}, pages 8527--8540. Association for Computational Linguistics.

\bibitem[{Feng et~al.(2023)Feng, Lu, Liu, Zhan, and Wu}]{FengLLZW23}
Yujie Feng, Zexin Lu, Bo~Liu, Liming Zhan, and Xiao{-}Ming Wu. 2023.
\newblock Towards llm-driven dialogue state tracking.
\newblock In \emph{Proceedings of the 2023 Conference on Empirical Methods in
  Natural Language Processing, {EMNLP} 2023, Singapore, December 6-10, 2023},
  pages 739--755. Association for Computational Linguistics.

\bibitem[{Fern{\'a}ndez et~al.(2025)Fern{\'a}ndez, Fern{\'a}ndez, and
  Aceta}]{20}
Cristina Fern{\'a}ndez, Izaskun Fern{\'a}ndez, and Cristina Aceta. 2025.
\newblock Lamia: An llm approach for task-oriented dialogue systems in industry
  5.0.
\newblock In \emph{Proceedings of the 15th International Workshop on Spoken
  Dialogue Systems Technology}, pages 205--214.

\bibitem[{He et~al.(2022)He, Dai, Zheng, Wu, Cao, Liu, Jiang, Yang, Huang, Si
  et~al.}]{17}
Wanwei He, Yinpei Dai, Yinhe Zheng, Yuchuan Wu, Zheng Cao, Dermot Liu, Peng
  Jiang, Min Yang, Fei Huang, Luo Si, and 1 others. 2022.
\newblock Galaxy: A generative pre-trained model for task-oriented dialog with
  semi-supervised learning and explicit policy injection.
\newblock In \emph{Proceedings of the AAAI conference on artificial
  intelligence}, volume~36, pages 10749--10757.

\bibitem[{Heck et~al.(2020)Heck, van Niekerk, Lubis, Geishauser, Lin, Moresi,
  and Gasic}]{HeckNLGLMG20}
Michael Heck, Carel van Niekerk, Nurul Lubis, Christian Geishauser,
  Hsien{-}Chin Lin, Marco Moresi, and Milica Gasic. 2020.
\newblock Trippy: {A} triple copy strategy for value independent neural dialog
  state tracking.
\newblock In \emph{Proceedings of the 21th Annual Meeting of the Special
  Interest Group on Discourse and Dialogue, SIGdial 2020, 1st virtual meeting,
  July 1-3, 2020}, pages 35--44. Association for Computational Linguistics.

\bibitem[{Kwan et~al.(2023)Kwan, Wang, Wang, and Wong}]{25}
Wai-Chung Kwan, Hong-Ru Wang, Hui-Min Wang, and Kam-Fai Wong. 2023.
\newblock A survey on recent advances and challenges in reinforcement learning
  methods for task-oriented dialogue policy learning.
\newblock \emph{Machine Intelligence Research}, 20(3):318--334.

\bibitem[{Lin et~al.(2021)Lin, Liu, Madotto, Moon, Zhou, Crook, Wang, Yu, Cho,
  Subba, and Fung}]{LinLMMZCWYCSF21}
Zhaojiang Lin, Bing Liu, Andrea Madotto, Seungwhan Moon, Zhenpeng Zhou, Paul~A.
  Crook, Zhiguang Wang, Zhou Yu, Eunjoon Cho, Rajen Subba, and Pascale Fung.
  2021.
\newblock Zero-shot dialogue state tracking via cross-task transfer.
\newblock In \emph{Proceedings of the 2021 Conference on Empirical Methods in
  Natural Language Processing, {EMNLP} 2021, Virtual Event / Punta Cana,
  Dominican Republic, 7-11 November, 2021}, pages 7890--7900. Association for
  Computational Linguistics.

\bibitem[{Nguyen et~al.(2025)Nguyen, Chieu, Pham, and Bui}]{nguyen2025spec}
Vinh~Quang Nguyen, Nguyen~Quang Chieu, Hoang~Viet Pham, and Khac-Hoai~Nam Bui.
  2025.
\newblock Spec-tod: A specialized instruction-tuned llm framework for efficient
  task-oriented dialogue systems.
\newblock In \emph{Proceedings of the 26th Annual Meeting of the Special
  Interest Group on Discourse and Dialogue}, pages 133--145.

\bibitem[{Pan et~al.(2023)Pan, Chen, Xu, Che, and Qin}]{abs-2304-04256}
Wenbo Pan, Qiguang Chen, Xiao Xu, Wanxiang Che, and Libo Qin. 2023.
\newblock \href {https://doi.org/10.48550/arXiv.2304.04256} {A preliminary
  evaluation of chatgpt for zero-shot dialogue understanding}.
\newblock \emph{CoRR}, abs/2304.04256.

\bibitem[{Peng et~al.(2017)Peng, Li, Li, Gao, Celikyilmaz, Lee, and
  Wong}]{PengLLGCLW17}
Baolin Peng, Xiujun Li, Lihong Li, Jianfeng Gao, Asli Celikyilmaz, Sungjin Lee,
  and Kam{-}Fai Wong. 2017.
\newblock Composite task-completion dialogue policy learning via hierarchical
  deep reinforcement learning.
\newblock In \emph{Proceedings of the 2017 Conference on Empirical Methods in
  Natural Language Processing, {EMNLP} 2017, Copenhagen, Denmark, September
  9-11, 2017}, pages 2231--2240. Association for Computational Linguistics.

\bibitem[{Qin et~al.(2023)Qin, Pan, Chen, Liao, Yu, Zhang, Che, and
  Li}]{0001PCLY0CL23}
Libo Qin, Wenbo Pan, Qiguang Chen, Lizi Liao, Zhou Yu, Yue Zhang, Wanxiang Che,
  and Min Li. 2023.
\newblock End-to-end task-oriented dialogue: {A} survey of tasks, methods, and
  future directions.
\newblock In \emph{Proceedings of the 2023 Conference on Empirical Methods in
  Natural Language Processing, {EMNLP} 2023, Singapore, December 6-10, 2023},
  pages 5925--5941. Association for Computational Linguistics.

\bibitem[{Reimers and Gurevych(2019)}]{36}
Nils Reimers and Iryna Gurevych. 2019.
\newblock Sentence-bert: Sentence embeddings using siamese bert-networks.
\newblock In \emph{Proceedings of the 2019 Conference on Empirical Methods in
  Natural Language Processing and the 9th International Joint Conference on
  Natural Language Processing, {EMNLP-IJCNLP} 2019, Hong Kong, China, November
  3-7, 2019}, pages 3980--3990. Association for Computational Linguistics.

\bibitem[{Rohmatillah and Chien(2023)}]{24}
Mahdin Rohmatillah and Jen-Tzung Chien. 2023.
\newblock Hierarchical reinforcement learning with guidance for multi-domain
  dialogue policy.
\newblock \emph{IEEE/ACM Transactions on Audio, Speech, and Language
  Processing}, 31:748--761.

\bibitem[{Rohmatillah et~al.(2023)Rohmatillah, Chien et~al.}]{27}
Mahdin Rohmatillah, Jen-Tzung Chien, and 1 others. 2023.
\newblock Advances and challenges in multi-domain task-oriented dialogue policy
  optimization.
\newblock \emph{APSIPA Transactions on Signal and Information Processing},
  12(1).

\bibitem[{Schulman et~al.(2017)Schulman, Wolski, Dhariwal, Radford, and
  Klimov}]{34}
John Schulman, Filip Wolski, Prafulla Dhariwal, Alec Radford, and Oleg Klimov.
  2017.
\newblock Proximal policy optimization algorithms.
\newblock \emph{CoRR}.

\bibitem[{Wei et~al.(2025)Wei, Shan, and Li}]{turn1academia20}
Yuan Wei, Xiaohan Shan, and Jianmin Li. 2025.
\newblock \href {https://arxiv.org/abs/2503.21807} {Lero: Llm-driven
  evolutionary framework with hybrid rewards and enhanced observation for
  multi-agent reinforcement learning}.
\newblock \emph{arXiv}.

\bibitem[{Wu et~al.(2019)Wu, Madotto, Hosseini{-}Asl, Xiong, Socher, and
  Fung}]{WuMHXSF19}
Chien{-}Sheng Wu, Andrea Madotto, Ehsan Hosseini{-}Asl, Caiming Xiong, Richard
  Socher, and Pascale Fung. 2019.
\newblock Transferable multi-domain state generator for task-oriented dialogue
  systems.
\newblock In \emph{Proceedings of the 57th Conference of the Association for
  Computational Linguistics, {ACL} 2019, Florence, Italy, July 28- August 2,
  2019, Volume 1: Long Papers}, pages 808--819. Association for Computational
  Linguistics.

\bibitem[{Yi et~al.(2024)Yi, Ouyang, Liu, Liao, Xu, and Shen}]{abs-2402-18013}
Zihao Yi, Jiarui Ouyang, Yuwen Liu, Tianhao Liao, Zhe Xu, and Ying Shen. 2024.
\newblock A survey on recent advances in llm-based multi-turn dialogue systems.
\newblock \emph{CoRR}, abs/2402.18013.

\bibitem[{Yu et~al.(2023)Yu, Chen, and Yu}]{18}
Xiao Yu, Maximillian Chen, and Zhou Yu. 2023.
\newblock Prompt-based monte-carlo tree search for goal-oriented dialogue
  policy planning.
\newblock \emph{arXiv preprint arXiv:2305.13660}.

\bibitem[{Zhang et~al.(2024)Zhang, Huang, Wu, Liu, Zheng, Dong, Shen, Dou,
  Zhao, Ye et~al.}]{12}
Ming Zhang, Caishuang Huang, Yilong Wu, Shichun Liu, Huiyuan Zheng, Yurui Dong,
  Yujiong Shen, Shihan Dou, Jun Zhao, Junjie Ye, and 1 others. 2024.
\newblock Transfertod: A generalizable chinese multi-domain task-oriented
  dialogue system with transfer capabilities.
\newblock \emph{arXiv preprint arXiv:2407.21693}.

\bibitem[{Zhao et~al.(2024)Zhao, Dastani, Long, Wang, and Wang}]{ZhaoDLWW24}
Yangyang Zhao, Mehdi Dastani, Jinchuan Long, Zhenyu Wang, and Shihan Wang.
  2024.
\newblock Rescue conversations from dead-ends: Efficient exploration for
  task-oriented dialogue policy optimization.
\newblock \emph{Trans. Assoc. Comput. Linguistics}, 12:1578--1596.

\bibitem[{Zhou et~al.(2024)Zhou, Liu, Dong, and Liu}]{28}
Zhenyou Zhou, Zhibin Liu, Zhaoan Dong, and Yuhan Liu. 2024.
\newblock Model discrepancy policy optimization for task-oriented dialogue.
\newblock \emph{Computer Speech \& Language}, 87:101636.

\bibitem[{Zhu et~al.(2020{\natexlab{a}})Zhu, Huang, Zhang, Zhu, and
  Huang}]{ZhuHZZH20}
Qi~Zhu, Kaili Huang, Zheng Zhang, Xiaoyan Zhu, and Minlie Huang.
  2020{\natexlab{a}}.
\newblock Crosswoz: {A} large-scale chinese cross-domain task-oriented dialogue
  dataset.
\newblock \emph{Trans. Assoc. Comput. Linguistics}, 8:281--295.

\bibitem[{Zhu et~al.(2020{\natexlab{b}})Zhu, Zhang, Fang, Li, Takanobu, Li,
  Peng, Gao, Zhu, and Huang}]{33}
Qi~Zhu, Zheng Zhang, Yan Fang, Xiang Li, Ryuichi Takanobu, Jinchao Li, Baolin
  Peng, Jianfeng Gao, Xiaoyan Zhu, and Minlie Huang. 2020{\natexlab{b}}.
\newblock Convlab-2: An open-source toolkit for building, evaluating, and
  diagnosing dialogue systems.
\newblock \emph{arXiv preprint arXiv:2002.04793}.

\end{thebibliography}


\begin{thebibliography}{0}
\providecommand{\natexlab}[1]{#1}

\end{thebibliography}

\appendix

\begin{table*}[th]
	\centering
	\caption{Respondent LLM CoT Prompt for Inferring Explicit and Implicit Constraints.}
	\label{tab:cot_prompt}
	\renewcommand{\arraystretch}{1.2}
	\scalebox{1}{
		\begin{tabular}{p{0.02\textwidth} p{0.9\textwidth}}
			\hline
			\textbf{Step} & \textbf{Prompt Instruction} \\
			\hline
			1 & \textbf{System Role (rea\_system)}: You are a helpful assistant. Focus only on the \texttt{belief\_state} of the user status. Fill in blank slots based on the known information across domains, without adding extra slots. Provide a confidence coefficient between 0 and 1 indicating your certainty. Domains refer to top-level keys in \texttt{belief\_state} (e.g., 'police', 'taxi', 'hotel', 'train'). \\[2mm]
			
			2 & \textbf{Main Prompt (rea\_main)}: Examples for step-by-step reasoning. \\[1mm]
			
			& \textbf{Example 1:} \\
			& User State: \texttt{\{user\_action: [], system\_action: [], belief\_state: \{...\}, request\_state: \{\}, terminated: False, history: []\}} \\
			& Step 1: Identify relevant task domains from the user status. \\
			& Step 2: Extract known slot information per domain. \\
			& Step 3: Analyze relationships and infer potential slot values logically (explicit and implicit). \\
			& Step 4: Assign a confidence coefficient to the inferred values. \\
			& Step 5: Produce final output in the format: @\{updated user status\}@, confidence coefficient: \$0.95\$. \\[1mm]
			
			& \textbf{Example 2:} Similar procedure with attention to local context inference and uncertainty handling. Confidence coefficient: \$0.87\$. \\[1mm]
			
			& \textbf{Example 3:} Inference with high ambiguity due to multiple plausible scenarios; confidence coefficient: \$0.65\$. \\[1mm]
			
			& \textbf{Example 4:} Extreme uncertainty in inference from limited information; confidence coefficient: \$0.35\$. \\[2mm]
			
			3 & \textbf{Instructions for Respondent LLM}:
			\begin{itemize}
				\item Ensure output format is consistent with input, enclosed with '@' at start and end.
				\item Include confidence coefficient (\$0-1\$) in the output, enclosed with '\$'.
				\item Do not add comments or extra slots, and do not modify \texttt{user\_action}, \texttt{system\_action}, \texttt{request\_state}, \texttt{terminated}, \texttt{history}.
				\item Focus solely on filling empty slot values.
			\end{itemize} \\[1mm]
			
			4 & \textbf{Question Template (question)}: Analyze and infer the information for the following user status:
			\begin{itemize}
				\item Q: \{user\_status\}
				\item Perform step-by-step reasoning to infer missing slot values.
			\end{itemize} \\
			\hline
		\end{tabular}
	}
\end{table*}

\section{Respondent LLM Prompt Design}
\label{app:A}

The Tab.~\ref{tab:cot_prompt} presents the step-by-step CoT prompt designed for the Respondent LLM to infer both explicit and implicit constraints in multi-domain TOD tasks.

\section{Prompt for LLM-only Policy Optimization}
\label{app:B}

In the \textbf{w/o RL (LLM-only)} ablation, the RL-based policy optimizer is removed and the LLM is directly prompted to select an action from the predefined action set. The Tab.~\ref{tab:llm_prompt} presents the exact prompt template used for Qwen-14B.

\section{Extendability to Different RL Backbones}
\label{appendix:rl-backbone}

VLK-RL enhances dialogue policies by augmenting the dialogue state with verified explicit and implicit cross-task constraints, while leaving the policy architecture, action space, and reward design unchanged.
This design suggests that the framework should be compatible with different reinforcement learning optimizers.
To empirically this claim
\clearpage

\begin{table*}[!t]
	\centering
	\caption{Prompt design for the LLM-only setting (w/o RL). The LLM is required to select one action $a \in \mathcal{A}$ based on the dialogue history and database results.}
	\label{tab:llm_prompt}
	\scalebox{0.95}{
		\begin{tabular}{p{0.95\textwidth}}
			\hline
			\textbf{System Instruction:} \\
			You are a task-oriented dialogue agent. Your goal is to select the next system action from the predefined action set $\mathcal{A}$ based on the current dialogue history and database state. \\
			\textbf{Action Set:} \\
			\{ \texttt{inform}\_{slot}, \texttt{request}\_{slot}, \texttt{confirm}\_{slot},  \texttt{book}, \texttt{goodbye},... \} \\
			\textbf{User Dialogue History:} \\
			Dialogue context $C_t$ up to current turn $t$ \\
			\textbf{Database Results:} \\
			Relevant slot–value information retrieved from DB \\
			\textbf{Task:} \\
			Based on the dialogue history and database results, select exactly one action from the action set. \\
			Do not generate free text or explanations. Only output the action name. \\
			\textbf{Output Format:} \\
			Action = [selected action] \\
			\hline
		\end{tabular}
	}
\end{table*}

\begin{table*}[!t]
	\centering
	\caption{Performance of VLK-RL with different RL backbones on MultiWOZ 2.1.}
	\label{tab:rl_backbone}
	\scalebox{0.7}{
		\begin{tabular}{lccccccc}
			\hline
			\textbf{Model} &
			\textbf{Avg. Precision} &
			\textbf{Avg. F1} &
			\textbf{Avg. Recall} &
			\textbf{Complete/Tot} &
			\textbf{Success/Tot} &
			\textbf{Avg. Turn (Succ)} &
			\textbf{Avg. Turn (All)} \\
			\hline
			DQN 
			& 0.5396 & 0.6750 & 0.7592 & 0.7093 & 0.3124 & 15.60 & 21.00 \\
			\cellcolor{blue!5}VLK-RL (Qwen-14B + DQN) 
			& \cellcolor{blue!5}0.6493 &\cellcolor{blue!5} 0.7050 &\cellcolor{blue!5} 0.8087 & \cellcolor{blue!5}0.7791 & \cellcolor{blue!5}0.6782 &\cellcolor{blue!5} 13.20 & \cellcolor{blue!5}18.50 \\
			\hline
			PG 
			& 0.5228 & 0.6511 & 0.7429 & 0.6947 & 0.2863 & 16.15 & 21.84 \\
			\cellcolor{blue!5}VLK-RL (Qwen-14B + PG) 
			&\cellcolor{blue!5} 0.6216 & \cellcolor{blue!5}0.6887 & \cellcolor{blue!5}0.7914 &\cellcolor{blue!5} 0.7579 & \cellcolor{blue!5}0.6418 & \cellcolor{blue!5}13.85 &\cellcolor{blue!5} 19.15 \\
			\hline
		\end{tabular}
	}
\end{table*}

\begin{table*}[!h]
	\centering
	\caption{Effect of Judge-Respondent model instantiation on VLK-RL performance on MultiWOZ 2.1.}
	\label{tab:cross_model}
	\scalebox{0.7}{
		\begin{tabular}{lccccccc}
			\hline
			\textbf{Judge / Respondent} &
			\textbf{Avg. Precision} &
			\textbf{Avg. F1} &
			\textbf{Avg. Recall} &
			\textbf{Complete/Tot} &
			\textbf{Success/Tot} &
			\textbf{Avg. Turn (Succ)} &
			\textbf{Avg. Turn (All)} \\
			\hline
			
			\cellcolor{blue!5}	Qwen-14B / Qwen-14B
			& \cellcolor{blue!5}\textbf{0.6628} & \cellcolor{blue!5}\textbf{0.7182} & \cellcolor{blue!5}\textbf{0.8429} & \cellcolor{blue!5}\textbf{0.8006} &\cellcolor{blue!5} \textbf{0.7214} &\cellcolor{blue!5} \textbf{12.51} & \cellcolor{blue!5}\textbf{17.35} \\

			GPT-4o-mini / Qwen-14B
			& 0.6481 & 0.7013 & 0.8315 & 0.7794 & 0.7026 & 12.84 & 17.68 \\
			
			Qwen-14B / GPT-4o-mini
			& 0.6517 & 0.7059 & 0.8352 & 0.7831 & 0.7089 & 12.77 & 17.59 \\
			\hline
		\end{tabular}
	}
\end{table*}

\begin{figure*}[!t]
	\includegraphics[width=2.0\columnwidth]{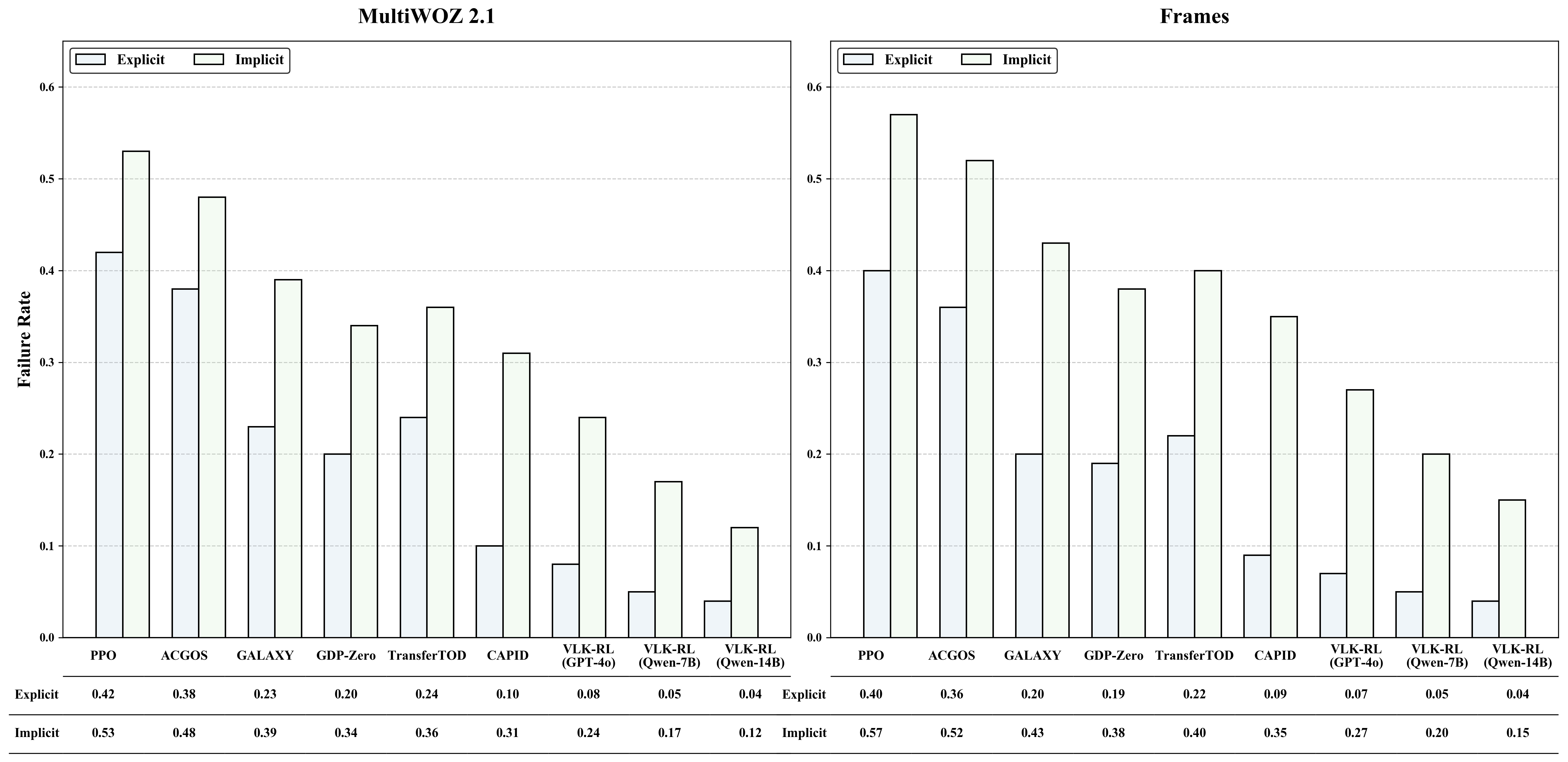}
	\caption{Failure rates for explicit and implicit constraints on MultiWOZ 2.1 and Frames separately. }
	\label{app:data-level}
\end{figure*}

\FloatBarrier

\begin{table*}[t]
	\centering
	\caption{Qualitative examples of LLM failures and the effect of cross-examination.}
	\label{tab:cross_exam_examples}
	\scalebox{0.95}{
		\begin{tabular}{p{3cm}p{6cm}p{6cm}}
			\hline
			\textbf{Failure Type} & \textbf{Before Mechanism} & \textbf{After Mechanism} \\
			\hline
			T1: Hallucinations &
			User: "I’d like a moderately priced restaurant in the centre of town."\par
			System: "You can try \textcolor{red}{‘King’s Garden Bistro,’ which is open from 8 am to 11 pm and serves Italian food}." & 
			User: "I’d like a moderately priced restaurant in the centre of town."\par
			System: "I recommend \textcolor{red}{‘Cotto,’ a moderately priced restaurant in the city centre} that matches your request." \\
			\hline
			T2: Inconsistencies &
			User: "I need a cheap guesthouse in the north of the city."\par
			System: "I found a cheap guesthouse in the north called ‘Acorn Guest House.’"\par
			User: "Can you remind me of the price range again?"\par
			System: "\textcolor{red}{‘Acorn Guest House’ is in the moderate price range and located in the centre}." &
			User: "I need a cheap guesthouse in the north of the city."\par
			System: "I found a cheap guesthouse in the north called ‘Acorn Guest House.’"\par
			User: "Can you remind me of the price range again?"\par
			System: "\textcolor{red}{‘Acorn Guest House’ is still in the cheap price range and located in the north}, as mentioned before." \\
			\hline
			T3: Constraint Violations &
			User: "Please book a vegetarian restaurant in the centre that is open after 9 PM."\par
			System: "I booked ‘The Garden’ for you. It’s vegetarian and in the centre, but \textcolor{red}{it closes at 8:30 PM}." & 
			User: "Please book a vegetarian restaurant in the centre that is open after 9 PM."\par
			System: "\textcolor{red}{‘The Garden’ closes at 8:30 PM and does not satisfy your time constraint}. Instead, I booked \textcolor{red}{‘Taste of Cambridge,’ which is open until 10 PM}." \\
			\hline
		\end{tabular}
	}
\end{table*}

\noindent we replace the default PPO optimizer with alternative RL algorithms and evaluate performance on the MultiWOZ 2.1 benchmark under the same experimental settings.
Specifically, we consider Deep Q-Networks (DQN) as a value-based method and vanilla Policy Gradient (PG) as a policy-based method.
All models are trained using identical dialogue state representations and hyperparameter tuning procedures, differing only in the RL backbone.

Tab.~\ref{tab:rl_backbone} reports the performance of VLK-RL instantiated with different reinforcement learning backbones on the MultiWOZ 2.1 dataset. For both value-based (DQN) and policy-based (PG) methods, incorporating VLK-RL consistently improves task success rate, dialogue completion rate, and state-level prediction metrics compared to their respective baselines.
Notably, while absolute performance varies across RL algorithms, the relative gains introduced by VLK-RL remain stable.
This indicates that the improvements primarily stem from enhanced constraint-aware state representations rather than optimizer-specific characteristics.
In addition, VLK-RL reduces the average number of dialogue turns required for successful task completion, suggesting improved dialogue efficiency across different optimization paradigms.

\section{Analysis of Judge-Respondent Model Instantiation}
\label{app:cross_model}

This appendix investigates whether the effectiveness of the proposed dual-role cross-examination mechanism
depends on using the same LLM for both the Judge and Respondent roles.
While the main experiments instantiate both roles with a unified LLM,
the core mechanism relies on role-specific prompting rather than architectural differences.
Using the same LLM ensures a shared logical framework and consistent knowledge boundaries,
which facilitates the detection of internal inconsistencies by reducing false positives
introduced by divergent reasoning styles across models. Although hallucinations may still occur, they tend to manifest differently across role-specific generations,
allowing the Judge to identify contradictions relative to its own reasoning trajectory.
In contrast, cross-model instantiations may introduce additional noise due to heterogeneous reasoning preferences,
leading to missed inconsistencies or incorrect rejections.
To empirically validate this claim, we conduct ablation experiments by cross-assigning GPT-4o-mini and Qwen-14B
to the Judge and Respondent roles, and compare them with the same-model configurations.
Qwen-7B is not included as it consistently underperforms Qwen-14B in our main experiments
and does not provide additional insight beyond model scale effects.

As shown in Tab.~\ref{tab:cross_model}, VLK-RL remains effective under both same-model and cross-model instantiations,
indicating that the proposed framework does not rely on a specific LLM pairing.
However, same-model configurations consistently yield slightly higher success rates and more stable dialogue efficiency,
supporting our choice of a unified LLM as the default setting. These results suggest that role-induced reasoning diversity, rather than model heterogeneity, is the primary driver of effective cross-examination.

\section{Constraint Failure Rates by Dataset}
\label{app:static}

Fig.~\ref{app:data-level} provides a dataset-level breakdown of constraint failure patterns, reporting the failure rates of explicit and implicit constraints on MultiWOZ 2.1 and Frames separately to complement the aggregated analysis in the main paper.

\section{Importance of Dual-role LLM Cross-Examination}
\label{app:cross_exam}

\subsection{Quantitative Analysis of LLM Failure Modes}
To justify the necessity of dual-role LLM cross-examination, we quantify hallucinations, inconsistencies, and constraint violations in MultiWOZ 2.1 dialogues. We sampled 200 user goals and manually labeled three major failure types:
\begin{itemize}
	\item \textbf{T1: Hallucinations} (fabricated entities or attributes)
	\item \textbf{T2: Inconsistencies} (contradictions in responses or with prior turns)
	\item \textbf{T3: Constraint Violations} (ignoring user or system constraints)
\end{itemize}

\begin{table}[h]
	\centering
	\caption{Proportion of dialogues affected by failure types with and without cross-examination.}
	\label{tab:cross_exam_quant}
		\scalebox{0.7}{
	\begin{tabular}{lcc}
		\hline
		\textbf{Failure Type} & \textbf{Before Mechanism} & \textbf{After Mechanism} \\
		\hline
		T1: Hallucinations & 23.5\% & 6.5\% \\
		T2: Inconsistencies & 22.0\% & 7.5\% \\
		T3: Constraint Violations & 17.0\% & 5.0\% \\
		\hline
	\end{tabular}
}
\end{table}

The proportion of dialogues affected by each failure type is shown in Tab.~\ref{tab:cross_exam_quant}. The results show that hallucinations, inconsistencies, and constraint violations are frequent and systemic in ToD tasks. Dual-role cross-examination significantly reduces all three failure categories, validating its necessity.

\subsection{Qualitative Examples of Knowledge Failures}

Tab.~\ref{tab:cross_exam_examples} illustrates representative dialogue failures before and after introducing dual-role cross-examination.

\subsection{Alternative Designs for Dual-role Verification}

\begin{figure}[t]
	\includegraphics[width=\columnwidth]{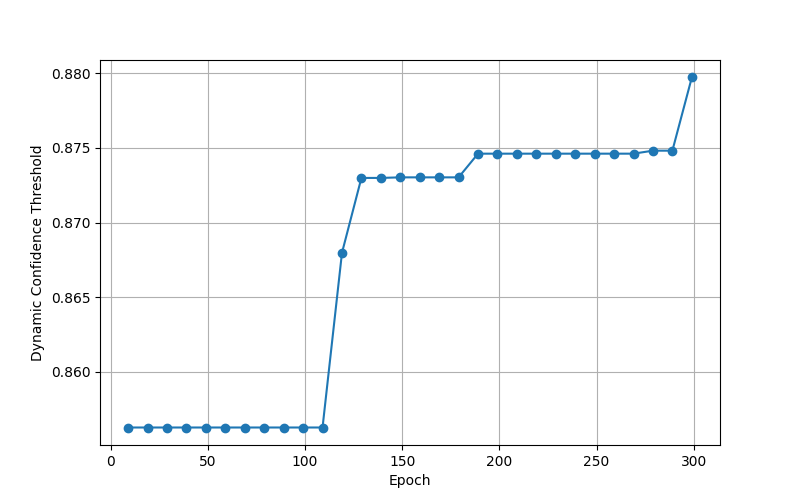}
	\caption{Dynamic changes in confidence threshold during training for VLK-RL (Qwen14b).
	}
	\label{epoch_trust}
\end{figure}

To examine whether dual-role cross-examination can be replaced by simpler single-model heuristics, we design two \emph{confidence-based gating} variants that rely on an LLM’s self-reported confidence to filter its own outputs. Unlike VLK-RL, which verifies knowledge via role-based debate, these variants accept or reject inferred knowledge without introducing an explicit Judge role. Specifically, we consider the following two alternatives:

\begin{itemize}
	\item \textbf{Confidence-fixed ($\tau = 0.85$).}  
	The LLM output is accepted only if its self-reported confidence exceeds a fixed threshold of $0.85$; otherwise, the inferred knowledge is discarded.
	
	\item \textbf{Confidence-dynamic.}  
	The acceptance threshold is adjusted during training according to validation performance. Formally, the threshold for epoch $e{+}1$ is updated as Eq.~\ref{eq:dynamic_threshold}.
\end{itemize}

\begin{equation}
	\label{eq:dynamic_threshold}
	\resizebox{0.9\hsize}{!}{$
	\tau^{(e+1)} =
	\begin{cases}
		\tau_{0} & \text{if } e < T_{\text{th}} \\[0.5em]
		\tau_{0} + \alpha \cdot \max(0, \text{F1}^{(e)} - \text{F1}_{\text{th}}) & \text{otherwise}
	\end{cases}
	$}
\end{equation}

\noindent where $\tau_{0}$ denotes the initial threshold, $\text{F1}^{(e)}$ is the validation F1 score at epoch $e$, $\text{F1}_{\text{th}}$ is the minimum target F1, and $T_{\text{th}}$ marks the training stage after which dynamic adjustment begins. Fig.~\ref{epoch_trust} illustrates the evolution of $\tau$ during training.

\begin{table*}[h]
	\centering
	\caption{Performance of alternative dual-role verification designs.}
	\label{tab:cross_exam_alternatives}
	\scalebox{0.7}{
		\begin{tabular}{lccccccc}
			\hline
			Model & Avg. Precision & Avg. F1 & Avg. Recall & Complete/Tot & Success/Tot & Avg. Turn (Succ) & Avg. Turn (All) \\
			\hline
			\cellcolor{blue!5}VLK-RL (Qwen-14B) & \cellcolor{blue!5}0.6628 & \cellcolor{blue!5}0.7182 &\cellcolor{blue!5} 0.8429 & \cellcolor{blue!5}0.8006 & \cellcolor{blue!5}0.7214 & \cellcolor{blue!5}12.51 &\cellcolor{blue!5} 17.35 \\
			~~~~w/o Cross-Examination & 0.4545 & 0.5252 & 0.7624 & 0.5400 & 0.4900 & 13.14 & 21.30 \\
			~~~~Confidence-fixed ($\tau=0.85$) & 0.6245 & 0.6784 & 0.8194 & 0.6100 & 0.6000 & 14.13 & 21.22 \\
			~~~~Confidence-dynamic & 0.6728 & 0.7243 & 0.8642 & 0.7100 & 0.6700 & 14.53 & 21.16 \\
			\hline
		\end{tabular}
	}
\end{table*}

\begin{table*}[t]
	\centering
	\caption{Ablation and prompt-based variants for the T2S mapper. Runtime is relative to VLK-RL (Qwen-14B).}
	\label{tab:t2s_alternatives}
	\scalebox{0.6}{
		\begin{tabular}{lcccccccc}
			\hline
			Model & Avg. Precision & Avg. F1 & Avg. Recall & Complete/Tot & Success/Tot & Avg. Turn (Succ) & Avg. Turn (All) & Runtime \\
			\hline
			\cellcolor{blue!5}VLK-RL (Qwen-14B) & \cellcolor{blue!5}0.6628 & \cellcolor{blue!5}0.7182 &\cellcolor{blue!5} 0.8429 & \cellcolor{blue!5}0.8006 & \cellcolor{blue!5}0.7214 & \cellcolor{blue!5}12.51 &\cellcolor{blue!5} 17.35 & \cellcolor{blue!5}1x \\
			~~~~w/o T2S Mapper & 0.4987 & 0.5623 & 0.7716 & 0.5732 & 0.5124 & 13.85 & 20.87 & 1x \\
			~~~~Mapper-prompt with retries (5x) & 0.5887 & 0.6415 & 0.8016 & 0.6503 & 0.5812 & 13.94 & 20.85 & 1.2x \\
			~~~~Mapper-prompt with retries (20x) & 0.6402 & 0.7056 & 0.8005 & 0.7882 & 0.7105 & 14.35 & 19.02 & 8.4x \\
			\hline
		\end{tabular}
	}
\end{table*}

Tab.~\ref{tab:cross_exam_alternatives} reports the performance of these alternatives. Both confidence-based gating strategies outperform naive removal of cross-examination, indicating that filtering unreliable LLM outputs is beneficial. However, neither variant matches the full VLK-RL framework. Although the dynamic threshold yields higher recall by gradually relaxing acceptance criteria, both gating strategies remain inferior to dual-role cross-examination. This gap arises because self-confidence estimates are systematically over-optimistic and cannot expose latent inconsistencies or constraint violations. In contrast, explicit role separation enables independent reasoning trajectories and more reliable detection of hallucinations, confirming the necessity of dual-role verification in VLK-RL.

\begin{table*}[]
	\centering
	\caption{Performance under low-resource environments, where all models are trained from scratch without pre-trained weights, warm-start initialization, or external data. Each epoch only stores one dialogue for training.}
	\label{tab:low-resource} 
	\scalebox{0.72}{
		\begin{tabular}{lccccccc}
			\hline
			\textbf{Model} & 
			\textbf{Avg. Precision} & 
			\textbf{Avg. F1} & 
			\textbf{Avg. Recall} & 
			\textbf{Complete/Tot} & 
			\textbf{Success/Tot} & 
			\textbf{Avg. Turn (Succ)} & 
			\textbf{Avg. Turn (All)} \\
			\hline
			PPO             & 0.1456 & 0.0935 & 0.0737 & 0.0312 & 0.0000 & --    & 28.45 \\
			ACGOS           & 0.1628 & 0.1154 & 0.0979 & 0.0421 & 0.0193 & 26.10 & 27.92 \\
			GALAXY          & 0.1875 & 0.1316 & 0.1164 & 0.0517 & 0.0326 & 24.83 & 27.35 \\
			GDP-Zero        & 0.2059 & 0.1587 & 0.1432 & 0.0562 & 0.0448 & 23.74 & 26.81 \\
			TransferTOD     & 0.1984 & 0.1492 & 0.1379 & 0.0541 & 0.0412 & 24.25 & 26.94 \\
			CAPID           & 0.2146 & 0.1659 & 0.1473 & 0.0618 & 0.0524 & 23.10 & 26.40 \\
			VLK-RL (GPT-4o-mini) & 0.2651 & 0.2053 & 0.1827 & 0.0912 & 0.0890 & 21.35 & 25.10 \\
			VLK-RL (Qwen-7B)     & 0.2789 & 0.2176 & 0.1941 & 0.0983 & 0.1034 & 20.62 & 24.71 \\
			\cellcolor{blue!5}VLK-RL (Qwen-14B)    & \cellcolor{blue!5}\textbf{0.2952} & \cellcolor{blue!5}\textbf{0.2351} & \cellcolor{blue!5}\textbf{0.2058} & \cellcolor{blue!5}\textbf{0.1105} & \cellcolor{blue!5}\textbf{0.1217} & \cellcolor{blue!5}\textbf{19.48} & \cellcolor{blue!5}\textbf{24.21} \\
			\hline
		\end{tabular}
	}
\end{table*}

\section{Importance of Ontology-aware Text-to-Slot Mapping}
\label{app:t2s_mapper}

To simulate direct LLM output without a dedicated mapper, we designed stricter prompts forcing slot-value formatting, \textit{Mapper-prompt with retries ($k$x)}. It denotes a prompt-only setting in which the LLM is instructed to directly output ontology-aligned slot-value pairs. After each generation attempt, the output is validated by a deterministic parser against the predefined ontology schema. If the output contains invalid slot names, malformed values, missing mandatory fields, or violates the expected key--value format, it is rejected and the model is re-prompted. This process repeats until either a valid structured output is obtained or the maximum number of retries $k$ is reached. If all attempts fail, the output for that turn is discarded. We consider two retry budgets:

\begin{itemize}
	\item \textbf{Mapper-prompt with retries (5x):} The LLM is allowed up to five regeneration attempts per dialogue turn. This setting balances computational cost and structural validity, but often discards semantically correct outputs due to formatting errors when the retry limit is reached.
	\item \textbf{Mapper-prompt with retries (20x):} The LLM is allowed up to twenty regeneration attempts per turn. This aggressive retry strategy substantially increases the probability of obtaining a valid structured output, at the expense of significantly higher latency and inference cost.
\end{itemize}

As shown in Tab.~\ref{tab:t2s_alternatives}, increasing the retry budget from 5x to 20x consistently improves task success and completion rates, indicating that brute-force regeneration can partially recover structural correctness. However, this improvement comes with diminishing returns and substantial overhead: the 20x variant incurs an approximately $8\times$ increase in runtime compared to the original VLK-RL pipeline, rendering it impractical for real-world deployment. Moreover, both retry-based variants exhibit higher dialogue length and instability, reflecting the inherent variance of prompt-based structured generation.

In summary, the dual-role cross-examination and T2S mapper are complementary: the cross-examination ensures correctness and consistency of generated knowledge, while the T2S mapper guarantees structured, database-compatible states. Together, they provide high-quality input for RL optimization, leading to robust multi-turn dialogue performance.

\section{Low-resource Environments}
\label{app:scratch}

We evaluate all models under an extreme low-resource setting on MultiWOZ 2.1, where agents are trained entirely from scratch without pre-trained weights, warm-start initialization, or external data.  
We focus on MultiWOZ in this setting because it provides sufficient domain diversity and interaction complexity to stress-test policy learning under sparse supervision, while remaining computationally feasible for repeated from-scratch training. In contrast, Frames contains fewer but more tightly coupled subtasks, where training instability dominates under such limited data, making reliable comparison difficult.

As shown in Tab.~\ref{tab:low-resource}, all methods suffer substantial performance degradation. Classical RL baselines (PPO, ACGOS) exhibit near-zero success rates, indicating their inability to explore effective cross-domain strategies without prior knowledge. LLM-based approaches (GALAXY, GDP-Zero, TransferTOD) perform slightly better, leveraging in-context reasoning, but still struggle to generate executable and consistent slot–value pairs, leading to low task success and long dialogues. DST methods (CAPID) further improve over LLM-only baselines by enhancing cross-domain state tracking; however, without explicit modeling of commonsense-driven feasibility constraints, its gains remain limited under sparse supervision.
In contrast, all VLK-RL variants achieve markedly higher success rates and shorter dialogues. Notably, VLK-RL (Qwen-14B) attains over 12\% success, demonstrating that verified constraint extraction and structured slot grounding effectively compensate for the lack of training data. These results highlight the robustness of VLK-RL: by explicitly grounding validated constraints at the state level, the framework enables more stable and data-efficient policy optimization even in severely low-resource environments.

\end{document}